\documentclass[runningheads]{llncs}

\usepackage{graphicx}
\usepackage{amsmath}
\usepackage{amssymb}
\usepackage{mathrsfs}
\usepackage{bbm}
\usepackage{booktabs}
\usepackage{tikz}
\usepackage{todonotes}
\usepackage{dirtree}
\usetikzlibrary{arrows,shapes.misc}
\usepackage[caption=false]{subfig}
\usepackage{multirow}
\usepackage[hyphens]{url}

\usepackage{enumitem}

\newcommand{\mtt}[1]{\text{\normalfont{\texttt{#1}}}}
\newcommand{\narytuplesset}{\mathcal{T}}
\newcommand{\ontology}{\mathcal{O}}
\newcommand{\ontologyd}{\mathcal{O}_\mtt{Drug}}
\newcommand{\ontologyp}{\mathcal{O}_\mtt{Phenotype}}
\newcommand{\kb}{\mathcal{K}}
\newcommand{\ci}{\mathrm{ci}}
\newcommand{\msc}{\mathrm{msc}}
\newcommand{\msci}{\mathrm{msci}}
\newcommand{\leqp}[1]{\preccurlyeq^{\mtt{#1}}}
\newcommand{\simp}[1]{\sim^{\mtt{#1}}}
\newcommand{\leqa}{\preccurlyeq^{\ontology}}
\newcommand{\sima}{\sim^{\ontology}}
\newcommand{\leqi}{\preccurlyeq_{i}}
\newcommand{\simi}{\sim_{i}}
\newcommand{\leqIk}{\preccurlyeq_{I_k}}
\newcommand{\leqd}{\preccurlyeq^{\ontologyd}}

\newcommand{\leqop}{\preccurlyeq^{\ontologyp}}

\DTsetlength{0.2em}{8pt}{0.2em}{0.4pt}{1.6pt}

\spnewtheorem{matching-rule}{Rule}{\bfseries}{\normalfont}

\begin{document}

\title{Knowledge-Based Matching of $n$-ary Tuples\thanks{Supported by the \textit{PractiKPharma} project, founded by the French National Research Agency (ANR) under Grant ANR15-CE23-0028, by the IDEX ``Lorraine Universit\'e d'Excellence" (15-IDEX-0004), and by the \textit{Snowball} Inria Associate Team.}
}

\titlerunning{Knowledge-Based Matching of $n$-ary Tuples}

\author{Pierre Monnin\orcidID{0000-0002-2017-8426} \and
	Miguel Couceiro\orcidID{0000-0003-2316-7623} \and
	Amedeo Napoli \and
	Adrien Coulet\orcidID{0000-0002-1466-062X}}

\authorrunning{P.\@ Monnin et al.}

\institute{Universit\'e de Lorraine, CNRS, Inria, LORIA, F-54000 Nancy, France
\email{\{pierre.monnin, miguel.couceiro, amedeo.napoli, adrien.coulet\}@loria.fr}}

\maketitle 

\begin{abstract}
An increasing number of data and knowledge sources are accessible by human and software agents in the expanding Semantic Web.
Sources may differ in granularity or completeness, and thus be complementary.
Consequently, they should be reconciled in order to unlock the full potential of their conjoint knowledge.
In particular, units should be matched within and across sources, and their level of relatedness should be classified into equivalent, more specific, or similar. 
This task is challenging since knowledge units can be heterogeneously represented in sources (\textit{e.g.}, in terms of vocabularies).
In this paper, we focus on matching $n$-ary tuples in a knowledge base with a rule-based methodology. 
To alleviate heterogeneity issues, we rely on domain knowledge expressed by ontologies.
We tested our method on the biomedical domain of pharmacogenomics by searching alignments among 50,435 $n$-ary tuples from four different real-world sources.
Results highlight noteworthy agreements and particularities within and across sources. 

\end{abstract}

\section{Introduction}

In the Semantic Web~\cite{berners2001}, data or knowledge sources often describe similar units but may differ in quality, completeness, granularity, and vocabularies. 
Unlocking the full potential of the knowledge that these sources conjointly express requires matching equivalent, more specific, or similar knowledge units within and across sources.
This matching process results in alignments that enable the reconciliation of these sources, \textit{i.e.}, the harmonization of their content~\cite{euzenat2013}.
Such a reconciliation then provides a consolidated view of a domain that is useful in many applications, \textit{e.g.}, in knowledge fusion and fact-checking.

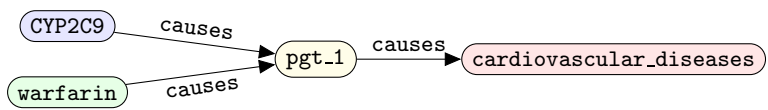
\begin{figure}
	\begin{center}
		\begin{tikzpicture}[scale=0.22,auto=center]
		\node[draw,rounded rectangle,fill=blue!10] (n1) at (-15,2) {\texttt{CYP2C9}};
		\node[draw,rounded rectangle,fill=green!10] (n2) at (-15,-2) {\texttt{warfarin}};
		\node[draw,rounded rectangle,fill=red!10] (n3) at (18,0) {\texttt{cardiovascular\_diseases}};
		\node[draw,rounded rectangle,fill=yellow!10] (n4) at (0,0) {\texttt{pgt\_1}};
		
		\draw[-triangle 45] (n1) -- (n4) node[midway,sloped,above] {\texttt{causes}};
		\draw[-triangle 45] (n2) -- (n4) node[midway,sloped,below] {\texttt{causes}};
		\draw[-triangle 45] (n4) -- (n3) node[midway,sloped,above] {\texttt{causes}};
		\end{tikzpicture}
	\end{center}
	\caption{Representation of a PGx relationship between gene \texttt{CYP2C9}, drug \texttt{warfarin} and phenotype \texttt{cardiovascular\_diseases}.
	It can be seen as an $n$-ary tuple $\mtt{pgt\_1} = \left(\left\{\mtt{warfarin}\right\}, \left\{\mtt{CYP2C9}\right\}, \left\{\mtt{cardiovascular\_diseases}\right\}\right)$.
	This tuple is reified through the individual \texttt{pgt\_1}, connecting its components through the \texttt{causes} predicate.}
	\label{figure:pgx-tuple}
\end{figure}

Here, we illustrate the interest of such a matching process to reconcile knowledge within the biomedical domain of pharmacogenomics (PGx), which studies the influence of genetic factors on drug response phenotypes.
PGx knowledge originates from distinct sources: reference databases such as PharmGKB, biomedical literature, or the mining of Electronic Health Records of hospitals.
Knowledge represented in these sources may differ in levels of validation, completeness, and granularity.
Consequently, reconciling these sources would provide a consolidated view on the knowledge of this domain, certainly beneficial in precision medicine, which aims at tailoring drug treatments to patients to reduce adverse effects and maximize drug efficacy~\cite{caudie,couletS16}.
PGx knowledge consists of $n$-ary relationships, here represented as tuples relating sets of drugs, sets of genomic variations, and sets of phenotypes.
Such an $n$-ary tuple states that a patient being treated with the specified sets of drugs, while having the specified genomic variations will be more likely to experience the given phenotypes, \textit{e.g.}, adverse effects.
For example, Figure~\ref{figure:pgx-tuple} depicts the tuple \mtt{pgt\_1}, which states that patients treated with warfarin may experience cardiovascular diseases because of variations in the CYP2C9 gene.
If a source contained the same tuple but with the genetic factor unknown, then it should be identified as less specific than \texttt{pgt\_1}.
Conversely, if a source contained the same tuple but with myocardial infarction as phenotype, then it should be identified as more specific than \mtt{pgt\_1}.

Motivated by this application, we propose a general and mathematically well-founded methodology to match $n$-ary tuples.
Precisely, given two $n$-ary tuples, we aim at deciding on their relatedness among five levels such as being equivalent or more specific.
We suppose that such tuples are represented within a knowledge base that is expressed using Semantic Web standards.
In such standards, only binary predicates exist, which requires the reification of $n$-ary tuples to represent them: tuples are individualized and linked to their components by predicates (see Figure~\ref{figure:pgx-tuple})~\cite{noy2006defining}.
In these knowledge bases, entities can also be associated with ontologies, \textit{i.e.}, formal representations of a domain~\cite{gruber1993translation}.
Ontologies consist of classes and predicates, partially ordered by the subsumption relation, denoted by $\sqsubseteq$.
This relation states that a class (respectively a predicate) is more specific than another. 

The process of matching $n$-ary tuples appears naturally in the scope of ontology matching~\cite{euzenat2013}, \textit{i.e.}, finding equivalences or subsumptions between classes, predicates, or instances of two ontologies. 
Here, we match individuals representing reified $n$-ary tuples, which is somewhat related to instance matching and the extraction of \textit{linkkeys}~\cite{atenciaDE14}.
However, we allow ourselves to state that a tuple is more specific than another, which is unusual in instance matching but common when matching classes or predicates with systems such as PARIS~\cite{suchanekAS11} and AMIE~\cite{galarragaPS13}.
Besides, to the best of our knowledge, works available in the literature do not deal with the complex task of matching $n$-ary tuples with potentially unknown arguments formed by sets of individuals. See Appendix~\ref{section:details-related-work} for further details.

In our approach, we assume that the tuples to match have the same arity, the same indices for their arguments, and that they are reified with the same predicates and classes.
Arguments are formed by sets of individuals (no literal values) and may be unknown.
This matching task thus reduces to comparing each argument of the tuples and aggregating these comparisons to establish their level of relatedness.
We achieve this process by defining five general rules, designed to satisfy some desired properties such as transitivity and symmetry.
To tackle the heterogeneity in the representation of tuples, we enrich this structure-based comparison with domain knowledge, \textit{e.g.}, the hierarchy of ontology classes and links between individuals.

This paper is organized as follows. 
In Section~\ref{section:problem-setting}, we formalize the problem of matching $n$-ary tuples. 
To tackle it, we propose two preorders in Section~\ref{section:preorders} to compare sets of individuals by considering domain knowledge: links between individuals, instantiations, and subsumptions.
These preorders are used in Section~\ref{section:matching-rules} to define matching rules that establish the level of relatedness between two $n$-ary tuples.
These rules are applied to PGx knowledge in Section~\ref{section:experimentation}. 
We discuss our results in Section~\ref{section:discussion} and present some directions of future work in Section~\ref{section:conclusion}.
Appendices are available online (\url{https://arxiv.org/abs/2002.08103}).

\section{Problem Setting}
\label{section:problem-setting}

We aim at matching $n$-ary tuples represented within a knowledge base $\kb$, \textit{i.e.}, we aim at determining the relatedness level of two tuples $t_1$ and $t_2$ (\textit{e.g.}, whether they are equivalent, more specific, or similar).
$\kb$ is represented in the formalism of Description Logics (DL)~\cite{baader2003} and thus consists of a TBox and an ABox.

Precisely, we consider a set $\narytuplesset$ of $n$-ary tuples to match.
This set is formed by tuples whose matching makes sense in a given application.
For example, in our use-case, $\narytuplesset$ consists of all PGx tuples from the considered sources.
All tuples in $\narytuplesset$ have the same arity $n$, and their arguments are sets of individuals of $\kb$.
Such a tuple $t$ can be formally represented as $t = \left(\pi_{1}(t), \dots, \pi_{n}(t) \right)$, where $\pi_{i}: \narytuplesset \to 2^\Delta$ is a mapping that associates each tuple $t$ to its $i$-th argument $\pi_{i}(t)$, which is a set of individuals included in the domain of interpretation $\Delta$.
The index set is the same for all tuples in $\narytuplesset$.
Tuples come from potentially noisy sources and some arguments may be missing.
As $\kb$ verifies the Open World Assumption, such arguments that are not explicitly specified as empty, can only be considered unknown and they are set to $\Delta$ to express the fact that all individuals may apply.
To illustrate, $\mtt{pgt\_1}$ in Figure~\ref{figure:pgx-tuple} could be seen as a ternary tuple $\mtt{pgt\_1} = \left(\left\{\mtt{warfarin}\right\}, \left\{\mtt{CYP2C9}\right\}, \left\{\mtt{cardiovascular\_diseases}\right\}\right)$, where arguments respectively represent the sets of involved drugs, genetic factors, and phenotypes.

In view of our formalism, matching two $n$-ary tuples $t_1$ and $t_2$ comes down to comparing their arguments $\pi_{i}(t_1)$ and $\pi_{i}(t_2)$ for each $i \in \left\{1, \dots, n \right\}$.
For instance, if $\pi_{i}(t_1) = \pi_{i}(t_2)$ for all $i$, then $t_1$ and $t_2$ are representing the same knowledge unit, highlighting an agreement between their sources.
In the next section, we propose other tests between arguments that are based on domain knowledge.

\section{Ontology-Based Preorders}
\label{section:preorders}

As previously illustrated, the matching of two $n$-ary tuples $t_1$ and $t_2$ relies on the comparison of each of their arguments $\pi_{i}(t_1)$ and $\pi_{i}(t_2)$, which are sets of individuals.
Such a comparison can be achieved by testing their inclusion or equality.
Thus, if $\pi_{i}(t_1) \subseteq \pi_{i}(t_2)$, then $\pi_{i}(t_1)$ can be considered as more specific than $\pi_{i}(t_2)$.
It is noteworthy that testing inclusion or equality implicitly considers \texttt{owl:sameAs} links that indicate identical individuals.
For example, the comparison of $\left\{\mtt{e}_1\right\}$ with $\left\{\mtt{e}_2\right\}$ while knowing that $\mtt{owl:sameAs}(\mtt{e}_1, \mtt{e}_2)$ results in an equality.
However, additional domain knowledge can be considered to help tackle the heterogeneous representation of tuples.
For instance, some individuals can be {\it part of} others. 
Individuals may also instantiate different ontological classes, which are themselves comparable through subsumption. 
To consider this domain knowledge in the matching process, we propose two preorders, {\it i.e.}, reflexive and transitive binary relations.

\subsection{Preorder $\leqp{p}$ Based on Links Between Individuals}

Several links may associate individuals in $\pi_{i}(t_j)$ with other individuals in $\kb$.
Some links involve a transitive and reflexive predicate (\textit{i.e.}, a preorder).
Then, for each such predicate \texttt{p}, we define a preorder $\leqp{p}$ parameterized by \texttt{p} as follows\footnote{See Appendix~\ref{section:proof-leqp} and Appendix~\ref{section:example-leqp} for the proof and examples.}:
\begin{equation}
\label{eq:leqp}
\pi_{i}(t_1) \leqp{p} \pi_{i}(t_2) \, \Leftrightarrow \, \forall e_1 \in \pi_{i}(t_1), \ \exists e_2 \in \pi_{i}(t_2),\ \kb \models \mtt{p}(e_1, e_2)
\end{equation}
Note that, from the reflexivity of \texttt{p} and the use of quantifiers $\forall$ and $\exists$, $\pi_{i}(t_1) \subseteq \pi_{i}(t_2)$ implies $\pi_{i}(t_1) \leqp{p} \pi_{i}(t_2)$.
The equivalence relation $\simp{p}$ associated with $\leqp{p}$ is defined as usual by:
\begin{equation}
\label{eq:simp}
\pi_{i}(t_1) \simp{p} \pi_{i}(t_2)\,  \Leftrightarrow \, \pi_{i}(t_1) \leqp{p} \pi_{i}(t_2) \text{ and } \pi_{i}(t_2) \leqp{p} \pi_{i}(t_1)
\end{equation}

\subsection{Preorder $\leqa$ Based on Instantiation and Subsumption}

The second preorder we propose takes into account classes of an ontology $\ontology$ ordered by subsumption and instantiated by individuals in $\pi_{i}(t_j)$.
We denote by $\mathrm{classes}(\ontology)$ the set of all classes of $\ontology$. 
As it is standard in DL, $\top$ denotes the largest class in $\ontology$.
Given an individual $e$, we denote by $\ci(\ontology, e)$ the set of classes of $\ontology$ instantiated by $e$ and distinct from $\top$, \textit{i.e.},
$$\ci(\ontology, e) = \left\{C \in \mathrm{classes}(\ontology) \backslash \left\{\top\right\} \mid \kb \models C(e) \right\}.$$
Note that $\ci(\ontology, e)$ may be empty. 
We explicitly exclude $\top$ from $\ci(\ontology, e)$ since $\kb$ may be incomplete.
Indeed, individuals may lack instantiations of specific classes but instantiate $\top$ by default.
Thus, $\top$ is excluded to prevent $\leqa$ from inadequately considering these individuals more general than individuals instantiating classes other than $\top$\footnote{See Appendix~\ref{section:exluding-top} for a detailed example.}.

Given $\mathcal{C} = \left\{C_1,\ C_2,\ \dots,\ C_k \right\} \subseteq \mathrm{classes}(\ontology)$, we denote by $\msc(\mathcal{C})$ the set of the most specific classes of $\mathcal{C}$, \textit{i.e.}, $\msc(\mathcal{C}) = \left\{C \in \mathcal{C} \mid \nexists D \in \mathcal{C},\ D \sqsubset C \right\}\footnote{$D \sqsubset C$ means that $D \sqsubseteq C$ and $D \not\equiv C$.}.$
Similarly, we denote by $\msci(\ontology, e)$ the set of the most specific classes of $\ontology$, except $\top$, instantiated by an individual $e$, \textit{i.e.}, $\msci(\ontology, e) = \msc(\ci(\ontology, e))$.

Given an  ontology $\ontology$, we define the preorder $\leqa$ based on set inclusion and subsumption as follows\footnote{See Appendix~\ref{section:proof-leqa} and Appendix~\ref{section:examples-leqa} for the proof and examples.}:
\begin{multline}
\label{eq:leqa}
\pi_{i}(t_1) \leqa \pi_{i}(t_2) \, \Leftrightarrow\,  \forall e_1 \in \pi_{i}(t_1), \ \Big[ \underbrace{e_1 \in \pi_{i}(t_2)}_\text{(\ref{eq:leqa}a)} \Big] \bigvee \Big[ \mathrm{msci}(\ontology, e_1) \neq \emptyset \ \wedge \\ \underbrace{\forall C_1 \in \mathrm{msci}(\ontology, e_1), \ 
\exists e_2 \in \pi_{i}(t_2),\ \exists C_2 \in \mathrm{msci}(\ontology, e_2),\ C_1 \sqsubseteq C_2}_{\text{(\ref{eq:leqa}b)}} \Big]
\end{multline}
Clearly, if $\pi_{i}(t_1)$ is more specific than $\pi_{i}(t_2)$ and $e_1 \in \pi_{i}(t_1)$, then (\ref{eq:leqa}a) $e_1 \in \pi_{i}(t_2)$, or (\ref{eq:leqa}b) all the most specific classes instantiated by $e_1$ are subsumed by at least one of the most specific classes instantiated by individuals in $\pi_{i}(t_2)$.
Thus individuals in $\pi_{i}(t_2)$ can be seen as ``more general'' than those in $\pi_{i}(t_1)$.
As before, $\leqa$ induces the equivalence relation $\sima$ defined by:
\begin{equation}
\label{eq:sima}
\pi_{i}(t_1) \sima \pi_{i}(t_2) \Leftrightarrow \pi_{i}(t_1) \leqa \pi_{i}(t_2) \text{ and } \pi_{i}(t_2) \leqa \pi_{i}(t_1)
\end{equation}
The preorder $\leqa$ can be seen as parameterized by the ontology $\ontology$, allowing to consider different parts of the TBox of $\kb$ for each argument $\pi_{i}(t_j)$, if needed. 

\section{Using Preorders to Define Matching Rules}
\label{section:matching-rules}

Let $t_1, t_2 \in \narytuplesset$ be two $n$-ary tuples to match.
We assume that each argument $i \in \left\{1, \dots, n\right\}$ is endowed with a preorder $\leqi \ \in \left\{\subseteq, \leqp{p}, \leqa\right\}$ that enables the comparison of $\pi_{i}(t_1)$ and $\pi_{i}(t_2)$.
We can define rules that aggregate such comparisons for all $i \in \left\{1, \dots, n\right\}$ and establish the relatedness level of $t_1$ and $t_2$.
Hence, our matching approach comes down to applying these rules to every ordered pair $(t_1, t_2)$ of $n$-ary tuples from $\narytuplesset$.

Here, we propose the following five relatedness levels: $=$, $\sim$, $\preccurlyeq$, $\lessgtr$, and $\propto$, from the strongest to the weakest.
Accordingly, we propose five matching rules of the form $B \Rightarrow H$, where $B$ expresses the conditions of the rule, testing equalities, equivalences, or inequalities between arguments of $t_1$ and $t_2$. 
Classically, these conditions can be combined using conjunctions or disjunctions, respectively denoted by $\wedge$ and $\vee$. 
If $B$ holds, $H$ expresses the relatedness between $t_1$ and $t_2$ to add to $\kb$ . 
Rules are applied from Rule~\ref{matching-rule:1} to Rule~\ref{matching-rule:5}.
Once conditions in $B$ hold for a rule, $H$ is added to $\kb$ and the following rules are discarded, meaning that at most one relatedness level is added to $\kb$ for each pair of tuples.
When no rule can be applied, $t_1$ and $t_2$ are considered incomparable and nothing is added to $\kb$.
The first four rules are the following:

\begin{matching-rule}
	\label{matching-rule:1}
	$\forall i \in \left\{1, \dots, n\right\}, \ \pi_{i}(t_1) = \pi_{i}(t_2) 
	\Rightarrow t_1 = t_2$
\end{matching-rule}

\begin{matching-rule}
	\label{matching-rule:2}
	$\forall i \in \left\{1, \dots, n\right\}, \ \pi_{i}(t_1) \simi \pi_{i}(t_2) 
	\Rightarrow t_1 \sim t_2$
\end{matching-rule}

\begin{matching-rule}
	\label{matching-rule:3}
	$\forall i \in \left\{1, \dots, n\right\}, \ \pi_{i}(t_1) \leqi \pi_{i}(t_2) 
	\Rightarrow t_1 \preccurlyeq t_2$
\end{matching-rule}

\begin{matching-rule}
	\label{matching-rule:4}
	$\forall i \in \left\{1, \dots, n\right\}, \ \left[\right.\left(\pi_{i}(t_1) = \pi_{i}(t_2)\right) \vee \left(\pi_{i}(t_2) \neq \Delta \wedge \pi_{i}(t_1) \leqi \pi_{i}(t_2) \right) \ \vee$ \\
	$\left(\pi_{i}(t_1) \neq \Delta \wedge \pi_{i}(t_2) \leqi \pi_{i}(t_1) \right) \left.\right] \Rightarrow t_1 \lessgtr t_2$
\end{matching-rule}

Rule~\ref{matching-rule:1} states that $t_1$ and $t_2$ are identical~($=$) whenever $t_1$ and $t_2$ coincide on each argument. 
Rule~\ref{matching-rule:2} states that $t_1$ and $t_2$ are equivalent~($\sim$) whenever each argument $i \in \left\{1, \dots, n \right\}$ of $t_1$ is equivalent to the same argument of $t_2$.
Rule~\ref{matching-rule:3} states that $t_1$ is more specific than $t_2$~($\preccurlyeq$) whenever each argument $i \in \left\{1, \dots, n \right\}$ of $t_1$ is more specific than the same argument of $t_2$ w.r.t. $\leqi$.
Rule~\ref{matching-rule:4} states that $t_1$ and $t_2$ have comparable arguments~($\lessgtr$) whenever they have the same specified arguments (\textit{i.e.}, different from  $\Delta$), and these arguments are comparable w.r.t. $\leqi$.
Rules~\ref{matching-rule:1} to~\ref{matching-rule:3} satisfy the transitivity property.
Additionally, Rules~\ref{matching-rule:1}, \ref{matching-rule:2}, and \ref{matching-rule:4} satisfy the symmetry property.

In Rules~\ref{matching-rule:1} to~\ref{matching-rule:4}, comparisons are made argument-wise. 
However, other relatedness cases may require to aggregate over arguments. 
For example, we may want to compare all individuals involved in two tuples, regardless of their arguments.
Alternatively, we may want to consider two tuples as weakly related if their arguments have a specified proportion of comparable individuals.
To this aim, we propose Rule~\ref{matching-rule:5}.
Let $\mathbb{I} = \left\{I_1, \dots, I_m\right\}$ be a partition of $\left\{1, \dots, n\right\}$, defined by the user at the beginning of the matching process.
We define the aggregated argument $I_k$ of $t_j$ as the union of all specified $\pi_{i}(t_j)$ (\textit{i.e.}, different from $\Delta$) for $i \in I_k$.
Formally, 
$$\pi_{I_k}(t_j) = \bigcup_{\substack{i \in I_k \\ \pi_{i}(t_j) \neq \Delta }}\pi_{i}(t_j).$$
We assume that each aggregated argument $I_k \in \mathbb{I}$ is endowed with a preorder $\leqIk \ \in \left\{\subseteq, \leqp{p}, \leqa\right\}$.
We denote by $\mathrm{SSD}(\pi_{I_k}(t_1), \pi_{I_k}(t_2))$ the semantic set difference between $\pi_{I_k}(t_1)$ and $\pi_{I_k}(t_2)$, \textit{i.e.}, $$\mathrm{SSD}(\pi_{I_k}(t_1), \pi_{I_k}(t_2)) = \left\{e_1 \mid e_1 \in \pi_{I_k}(t_1) \text{ and } \left\{e_1\right\} \not\leqIk \pi_{I_k}(t_2) \right\}.$$
Intuitively, it is the set of elements in $\pi_{I_k}(t_1)$ preventing it from being more specific than $\pi_{I_k}(t_2)$ w.r.t. $\leqIk$.
We define the operator $\propto_{I_k}$ as follows:
\[\pi_{I_k}(t_1) \propto_{I_k} \pi_{I_k}(t_2) = \begin{cases}
1 \text{ if } \pi_{I_k}(t_1) \leqIk \pi_{I_k}(t_2) \text{ or } \pi_{I_k}(t_2) \leqIk \pi_{I_k}(t_1) \\
1 - \frac{|\mathrm{SSD}(\pi_{I_k}(t_1), \pi_{I_k}(t_2)) \ \cup \ \mathrm{SSD}(\pi_{I_k}(t_2), \pi_{I_k}(t_1))|}{|\pi_{I_k}(t_1) \ \cup \ \pi_{I_k}(t_2)|} \text{ otherwise}
\end{cases} \]
\noindent This operator returns a number measuring the similarity between $\pi_{I_k}(t_1)$ and $\pi_{I_k}(t_2)$.
This number is equal to 1 if the two aggregated arguments are comparable.
Otherwise, it is equal to 1 minus the proportion of incomparable elements.
We denote by $\mathbb{I}_{\neq \Delta}(t_1, t_2) = \left\{I_k \mid I_k \in \mathbb{I} \text{ and } \pi_{I_k}(t_1) \neq \Delta \text{ and } \pi_{I_k}(t_2) \neq \Delta \right\}$ the set of aggregated arguments that are specified for both $t_1$ and $t_2$ (\textit{i.e.}, different from $\Delta$).
Then, Rule~\ref{matching-rule:5} is defined as follows:
\begin{matching-rule}
	\label{matching-rule:5}
	Let $\mathbb{I} = \left\{I_1, \dots, I_m\right\}$ be a partition of $\left\{1, \dots, n\right\}$, and let $\gamma_{\neq\Delta}$, $\gamma_S$, and $\gamma_C$ be three parameters, all fixed at the beginning of the matching process.
	\begin{multline*}
	\Big(|\mathbb{I}_{\neq \Delta}(t_1, t_2)| \ \geq \gamma_{\neq\Delta} \Big)
	\bigwedge \Big(\big[\forall I_k \in \mathbb{I}_{\neq \Delta}(t_1, t_2), \ \pi_{I_k}(t_1) \propto_{I_k} \pi_{I_k}(t_2) \geq \gamma_S\big] \ \vee \ \\
	\big[\big(\sum_{I_k \in \mathbb{I}_{\neq \Delta}(t_1, t_2)} \mathbbm{1}\left(\pi_{I_k}(t_1) \propto_{I_k} \pi_{I_k}(t_2) = 1\right)\big) \geq \gamma_C \big] \Big) \Rightarrow t_1 \propto t_2
	\end{multline*}
\end{matching-rule}
\noindent Rule~\ref{matching-rule:5} is applicable if at least $\gamma_{\neq\Delta}$ aggregated arguments are specified for both $t_1$ and $t_2$.
Then, $t_1$ and $t_2$ are weakly related~($\propto$) whenever all these specified aggregated arguments have a similarity of at least $\gamma_S$ or when at least $\gamma_C$ of them are comparable.
Notice that $\propto$ is symmetric.

\section{Application to Pharmacogenomic Knowledge}
\label{section:experimentation}

Our methodology was motivated by the problem of matching pharmacogenomic (PGx) tuples.
Accordingly, we tested this methodology on PGxLOD\footnote{\url{https://pgxlod.loria.fr}}~\cite{monnin2018pgxo}, a knowledge base represented in the $\mathcal{ALHI}$ Description Logic~\cite{baader2003}.
In PGxLOD, 50,435 PGx tuples were integrated from four different sources: (i) 3,650 tuples from structured data of PharmGKB, (ii) 10,240 tuples from textual portions of PharmGKB called clinical annotations, (iii) 36,535 tuples from biomedical literature, and (iv) 10 tuples from results found in EHR studies.
We obtained the matching results summarized in Table~\ref{table:results} and discussed in Section~\ref{section:discussion}. 
Details about formalization, code and parameters are given in  Appendix~\ref{section:details-experimentation}.

\begin{table}[t]
	\caption{Number of links resulting from each rule. Links are generated between tuples of distinct sources or within the same source.
	PGKB stands for ``PharmGKB'', sd for ``structured data'', and ca for ``clinical annotations''.
	As Rules~\ref{matching-rule:1}, \ref{matching-rule:2}, \ref{matching-rule:4}, and \ref{matching-rule:5} satisfy symmetry, links from $t_1$ to $t_2$ as well as from $t_2$ to $t_1$ are counted.
	Similarly, as Rules~\ref{matching-rule:1} to \ref{matching-rule:3} satisfy transitivity, transitivity-induced links are counted.
	Regarding \texttt{skos:broadMatch} links, rows represent origins and columns represent destinations.
}
	\label{table:results}
	\begin{center}
	\begin{tabular}{crcccc}        
		\toprule
	    & & PGKB (sd) & PGKB (ca) & Literature & EHRs \\
		\midrule
		\multirow{4}{3cm}{\centering Links from Rule~\ref{matching-rule:1}\\ Encoded by \texttt{owl:sameAs}} & PGKB (sd) & 166 & 0 & 0 & 0 \\
		& PGKB (ca) & 0 & 10,134 & 0 & 0 \\
		& Literature & 0 & 0 & 122,646 & 0\\
		& EHRs & 0 & 0 & 0 & 0 \\
		\midrule
		\multirow{4}{3cm}{\centering Links from Rule~\ref{matching-rule:2}\\Encoded by \texttt{skos:closeMatch}} & PGKB (sd) & 0 & 5 & 0 & 0 \\
		& PGKB (ca) & 5 & 1,366 & 0 & 0 \\
		& Literature & 0 & 0 & 16,692 & 0\\
		& EHRs & 0 & 0 & 0 & 0\\
		\midrule
		\multirow{4}{3cm}{\centering Links from Rule~\ref{matching-rule:3}\\Encoded by \texttt{skos:broadMatch}} & PGKB (sd) & 87 & 3 & 15 & 0 \\
		& PGKB (ca) & 9,325 & 605 & 42 & 0 \\
		& Literature & 0 & 0 & 75,138 & 0 \\
		& EHRs & 0 & 0 & 0 & 0\\
		\midrule
		\multirow{4}{3cm}{\centering Links from Rule~\ref{matching-rule:4} \\ Encoded by \texttt{skos:relatedMatch}} & PGKB (sd) & 20 & 0 & 0 & 0 \\
		& PGKB (ca) & 0 & 110 & 0 & 0 \\
		& Literature & 0 & 0 & 18,050 & 0\\
		& EHRs & 0 & 0 & 0 & 0\\
		\midrule
		\multirow{4}{3cm}{\centering Links from Rule~\ref{matching-rule:5}\\Encoded by \texttt{skos:related}} & PGKB (sd) & 100,596 & 287,670 & 414 & 2 \\
		& PGKB (ca) & 287,670 & 706,270 & 1,103 & 19 \\
		& Literature & 414 & 1,103 & 1,082,074 & 15 \\
		& EHRs & 2 & 19 & 15 & 0\\
		\bottomrule
	\end{tabular}
	\end{center}
\end{table}

\section{Discussion}
\label{section:discussion}

In Table~\ref{table:results}, we observe only a few inter-source links, which may be caused by missing mappings between the vocabularies used in sources.
Indeed, our matching process requires these mappings to compare individuals represented with different vocabularies.
This result underlines the relevance of enriching the knowledge base with ontology-to-ontology mappings.
We also notice that Rule~\ref{matching-rule:5} generates more links than the other rules, which emphasizes the importance of weaker relatedness levels to align sources and overcome their heterogeneity.
Some results were expected and therefore seem to validate our approach.
For example, some tuples from the literature appear more general than those of PharmGKB (with 15 and 42 \texttt{skos:broad\-Match} links).
These links are a foreseen consequence of the completion process of PharmGKB.
Indeed, curators achieve this completion after a literature review, 
inevitably leading to tuples more specific or equivalent to the ones in reviewed articles.
Interestingly, our methodology could ease such a review by pointing out articles describing similar tuples.
Clinical annotations of PharmGKB are in several cases more specific than structured data (9,325 \texttt{skos:broadMatch} links).
This is also expected as structured data are a broad-level summary of more complex phenotypes detailed in clinical annotations.

Regarding our method, using rules is somehow off the current machine learning trend~\cite{alamRMGR17a,nickelMTG15,ristoskiP16}. 
However, writing simple and well-founded rules constitutes a valid first step before applying machine learning approaches.
Indeed, such explicit rules enable generating a ``silver'' standard for matching, which may be useful to either train or evaluate supervised approaches.
Rules are readable and may thus be analyzed and confirmed by domain experts, and provide a basis of explanation for the matching results.
Additionally, our rules are simple enough to be generally true and useful in other domains.
By relying on instantiated classes and links between individuals, we illustrate how domain knowledge and reasoning mechanisms can serve a structure-based matching.
In future works, conditions under which preorders $\leqp{p}$ and $\leqa$ could be merged into one unique preorder deserve a deeper study.
See Appendix~\ref{section:details-discussion} for further discussion.

\section{Conclusion}
\label{section:conclusion}

In this paper, we proposed a rule-based approach to establish the relatedness level of $n$-ary tuples among five proposed levels.
It relies on rules and preorders that leverage domain knowledge and reasoning capabilities.
We applied our methodology to the real-world use case of matching pharmacogenomic relationships, and obtained insightful results.
In the future, we intend to compare and integrate our purely symbolic approach with ML methodologies.

\bibliographystyle{splncs04}
\bibliography{bibliography}

\newpage

\appendix

\section{Details about Related Works}
\label{section:details-related-work}

In ontology matching~\cite{euzenat2013}, existing works use different features to suggest alignments.
For example, some methods rely on the syntax of units label (\textit{e.g.}, string matching).
However, labels may not always be available.
Structure-based techniques can alleviate such limitations.
They rely either on the internal structure of a unit (\textit{i.e.}, predicates used to link a unit to literals) or the relational structure of a unit (\textit{i.e.}, its links with other units). 
Two examples of frequently considered relational structures are the hierarchy of classes and \texttt{partOf} links in an ontology.
As our matching approach compares tuples based on the individuals involved in their arguments and their associated domain knowledge, it relies on the relational structure of reified tuples. 

\textit{Linkkeys} are defined by Atencia \textit{et al.}~\cite{atenciaDE14} as a structure-based method to align individuals.
A linkkey consists of a pair of classes and a set of pairs of properties from two ontologies. 
Instances of these classes that share common values for all properties in the linkkey are regarded as identical.
Alternatively, PARIS is a holistic method proposed by Suchanek \textit{et al.}~\cite{suchanekAS11} to align individuals, classes, and predicates.
In this framework, alignments for each type of unit fertilize the others: they  are performed repeatedly until convergence. 
Rules in PARIS rely on the internal and relational structures of units and the functionality of predicates. 
Similarly, Galarraga \textit{et al.}~\cite{galarragaPS13} mine specific rules to align ontologies, using the AMIE system.
This system relies on the Partial Completeness Assumption, which is also built upon the functionality of predicates.

\section{Proof that $\leqp{p}$ is a preorder}
\label{section:proof-leqp}

\begin{proof} From the fact that \texttt{p} is reflexive, it immediately follows that  $\leqp{p}$ is reflexive. 
	Indeed, for every $E\subseteq \Delta$,  $E\leqp{p} E$ since for every $e\in E$, $\mtt{p}(e, e)$.
	
	To prove that $\leqp{p}$ is a preorder, it remains to show that $\leqp{p}$ is transitive.
	Consider $E_1, E_2,E_3 \subseteq \Delta$ such that:
	\[ E_1 \leqp{p} E_2 \  \text{and} \  E_2 \leqp{p} E_3.\] 
	In other words, $\forall e_1 \in E_1, \ \exists e_2 \in E_2,\ \kb \models \mtt{p}(e_1, e_2)$ and $ \forall e_2 \in E_2, \ \exists e_3 \in E_3,\ \kb \models \mtt{p}(e_2, e_3).$
	By the transitivity of \texttt{p}, we then have that  
	\[\forall e_1 \in E_1,  \ \exists e_3 \in E_3,\ \kb \models \mtt{p}(e_1, e_3),\]
	{\it i.e.}, $E_1 \leqp{p} E_3$. This shows that $\leqp{p}$ is transitive, and the proof is complete. \qed
\end{proof}

\section{Example of Use of $\leqp{p}$}
\label{section:example-leqp}

\begin{example}
	\texttt{partOf} is transitive and reflexive. 
	Thus, this predicate is a suitable candidate for the $\leqp{p}$ preorder.
	Consider three individuals $e_1$, $e_2,$ $e_3$ such that $\kb \models \mtt{partOf}(e_3, e_1)$. Then it follows that: 
	\begin{itemize}
		\item $\left\{e_1\right\} \leqp{partOf} \left\{e_1, e_2\right\}$, from set inclusion.
		\item $\left\{e_3, e_2 \right\} \leqp{partOf} \left\{e_1, e_2\right\}$.
		\item $\left\{e_3\right\}\leqp{partOf} \left\{e_1, e_2\right\}$.
		\item $\left\{e_3, e_1\right\}$ $\simp{partOf} \left\{e_1\right\}$. 
		As $e_3$ is a part of $e_1$, having both $e_3$ and $e_1$ in the same set can be seen as a redundancy. Such a case may arise in $\kb$ due to source heterogeneity. This redundancy is adequately identified by this equivalence result.
	\end{itemize}
\end{example}

\section{Details about Excluding $\top$ in $\ci(\ontology, e)$}
\label{section:exluding-top}

\begin{example}\label{example:2} Consider two PGx tuples $pgt_1$ and $pgt_2$ that involve the same drug and genetic factor.
Regarding the phenotype, $pgt_1$ is linked with an individual representing \textit{headache} that does not instantiate the class \textit{Headache} in $\ontology$ (\textit{e.g.}, MeSH) but instantiates $\top$ by default.
$pgt_2$ is linked with an individual \textit{pain} that instantiates $Pain$, with $Headache \sqsubseteq Pain$.
Intuitively, the knowledge expressed by $pgt_1$ is more specific than $pgt_2$.
However, by considering instantiated classes and knowing that $Pain \sqsubseteq \top$, $\leqa$ would inadequately conclude that $pgt_1$ is more general than $pgt_2$.
By excluding $\top$ from $\ci(\ontology, e)$, the tuples are incomparable, which avoids this unwanted behavior.
\end{example}

\section{Proof that $\leqa$ is a preorder}
\label{section:proof-leqa}

Recall that, for every $E_1, E_2 \subseteq \Delta$,
\begin{multline}
\label{eq:leqs}
E_1 \leqa E_2 \Leftrightarrow \forall e_1 \in E_1, \ \Big[ \underbrace{e_1 \in E_2}_{\text{(\ref{eq:leqs}a)}} \Big] \bigvee \Big[ \mathrm{msci}(\ontology, e_1) \neq \emptyset \ \wedge \\ \underbrace{\forall C_1 \in \mathrm{msci}(\ontology, e_1), \ 
	\exists e_2 \in E_2,\ \exists C_2 \in \mathrm{msci}(\ontology, e_2),\ C_1 \sqsubseteq C_2}_{\text{(\ref{eq:leqs}b)}} \Big]
\end{multline}

\begin{proof} The reflexivity of $\leqa$ follows immediately from (\ref{eq:leqs}a). 
	To see that it is also transitive, consider distinct $E_1, E_2, E_3 \subseteq \Delta$ such that $E_1 \leqa E_2 \  \text{and} \  E_2 \leqa E_3$. 
	We need to prove that $E_1 \leqa E_3$, that is,  
	\begin{multline*}
	\forall e_1 \in E_1, \ \Big[ e_1 \in E_3 \Big] \bigvee \Big[ \mathrm{msci}(\ontology, e_1) \neq \emptyset \ \wedge \\ \forall C_1 \in \mathrm{msci}(\ontology, e_1), \ 
	\exists e_3 \in E_3,\ \exists C_3 \in \mathrm{msci}(\ontology, e_3),\ C_1 \sqsubseteq C_3 \Big]
	\end{multline*}
	So let $e_1 \in E_1$. If $e_1 \in E_2$, then it follows from $E_2 \leqa E_3$ that
	\begin{multline}
	\label{eq:leqs-transitivity-1}
	\Big[ e_1 \in E_3 \Big] \bigvee \Big[ \mathrm{msci}(\ontology, e_1) \neq \emptyset \ \wedge \\ \forall C_1 \in \mathrm{msci}(\ontology, e_1), \ 
	\exists e_3 \in E_3,\ \exists C_3 \in \mathrm{msci}(\ontology, e_3),\ C_1 \sqsubseteq C_3\Big],
	\end{multline}
	and we are done. Otherwise, 
	\[\mathrm{msci}(\ontology, e_1) \neq \emptyset \ \wedge \ \forall C_1 \in \mathrm{msci}(\ontology, e_1),\ \exists e_2 \in E_2,\ \exists C_2 \in \mathrm{msci}(\ontology, e_2),\ C_1 \sqsubseteq C_2.\]
	As $E_2 \leqa E_3$, we have two possible cases for each $e_2 \in E_2$: 
	\begin{itemize}
		\item $e_2 \in E_3$ and for each $C_1 \in \mathrm{msci}(\ontology, e_1)$ we also have:
		\[\exists e_3 \in E_3,\ \exists C_3 \in \mathrm{msci}(\ontology, e_3),\ C_1 \sqsubseteq C_3, \ \text{or}\]
		\item $\exists e_3 \in E_3,\ \exists C_3 \in \mathrm{msci}(\ontology, e_3),\ C_2 \sqsubseteq C_3$. Since the  
		subsumption relation is transitive, $C_1 \sqsubseteq C_3$, and
		\[\exists e_3 \in E_3,\ \exists C_3 \in \mathrm{msci}(\ontology, e_3),\ C_1 \sqsubseteq C_3\]
	\end{itemize}
	From these two cases, it follows that for each $e_1 \in E_1$ such that
	\[\mathrm{msci}(\ontology, e_1) \neq \emptyset \ \wedge \ \forall C_1 \in \mathrm{msci}(\ontology, e_1),\ \exists e_2 \in E_2,\ \exists C_2 \in \mathrm{msci}(\ontology, e_2),\ C_1 \sqsubseteq C_2,\]
	\noindent we have that
	\begin{equation}
	\label{eq:leqs-transitivity-2}
	\exists e_3 \in E_3,\ \exists C_3 \in \mathrm{msci}(\ontology, e_3),\ C_1 \sqsubseteq C_3.
	\end{equation}
	
	From Equations~(\ref{eq:leqs-transitivity-1}) and (\ref{eq:leqs-transitivity-2}), it then follows that:
	\begin{multline*}
	\forall e_1 \in E_1, \ \Big[ e_1 \in E_3 \Big] \bigvee \Big[ \mathrm{msci}(\ontology, e_1) \neq \emptyset \ \wedge \\ \forall C_1 \in \mathrm{msci}(\ontology, e_1), \ 
	\exists e_3 \in E_3,\ \exists C_3 \in \mathrm{msci}(\ontology, e_3),\ C_1 \sqsubseteq C_3 \Big],
	\end{multline*}
	thus showing that $E_1 \leqa E_3$. As the latter holds for every $E_1, E_2, E_3 \subseteq \Delta$, $\leqa$ is transitive.
	\qed
\end{proof}

\section{Examples of Use of $\leqa$}
\label{section:examples-leqa}

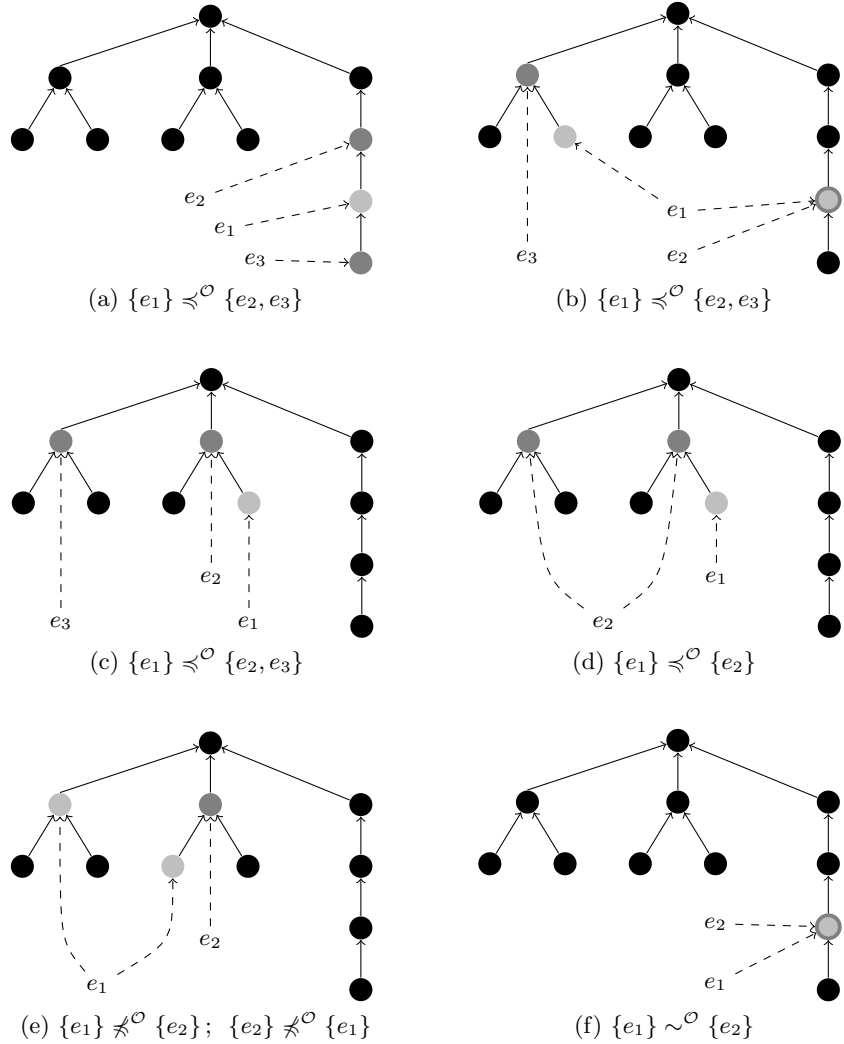
\begin{figure}
	\captionsetup[subfigure]{justification=centering,singlelinecheck=false}
	\begin{center}
		\subfloat[][$\left\{e_1\right\} \leqa \left\{e_2, e_3\right\}$]{\centering
			\begin{tikzpicture}[
			oclass/.style={circle, draw=none, fill=black,
				text centered, anchor=north, text=white},
			oclassD1/.style={circle, draw=none, fill=lightgray,
				text centered, anchor=north, text=white},
			oclassD2/.style={circle, draw=none, fill=gray,
				text centered, anchor=north, text=white},
			oclassD1D2/.style={circle, draw=gray, line width=0.5mm, fill=lightgray,
				text centered, anchor=north, text=white},
			level distance=0.5cm, growth parent anchor=south
			]
			\node (oc1) [oclass] {} [<-, sibling distance=2cm]
			child{ [sibling distance=1cm]
				node (oc2) [oclass] {}
				child{
					node (oc5) [oclass] {}
				}
				child { 
					node (oc6) [oclass] {}
				}
			}
			child { [sibling distance=1cm]
				node (oc4) [oclass] {}
				child{
					node (oc10) [oclass] {}
				}
				child { 
					node (oc11) [oclass] {}
				}
			}
			child { 
				node (oc3) [oclass] {}
				child{
					node (oc7) [oclassD2] {}
					child{
						node (oc8) [oclassD1] {}
						child{
							node (oc9) [oclassD2] {}
						}
					}
				}
			}
			;
			
			\node at (0.2, -3.0) (e1) {$e_1$};
			\node at (-0.2, -2.6) (e2) {$e_2$};
			\node at (0.6, -3.4) (e3) {$e_3$};
			
			\draw[dashed, ->] (e1) -- (oc8);
			\draw[dashed, ->] (e2) -- (oc7);
			\draw[dashed, ->] (e3) -- (oc9);
			\end{tikzpicture}
		}
		\qquad
		\qquad
		\subfloat[][$\left\{e_1\right\} \leqa \left\{e_2, e_3\right\}$]{\centering
			\begin{tikzpicture}[
			oclass/.style={circle, draw=none, fill=black,
				text centered, anchor=north, text=white},
			oclassD1/.style={circle, draw=none, fill=lightgray,
				text centered, anchor=north, text=white},
			oclassD2/.style={circle, draw=none, fill=gray,
				text centered, anchor=north, text=white},
			oclassD1D2/.style={circle, draw=gray, line width=0.5mm, fill=lightgray,
				text centered, anchor=north, text=white},
			level distance=0.5cm, growth parent anchor=south
			]
			\node (oc1) [oclass] {} [<-, sibling distance=2cm]
			child{ [sibling distance=1cm]
				node (oc2) [oclassD2] {}
				child{
					node (oc5) [oclass] {}
				}
				child { 
					node (oc6) [oclassD1] {}
				}
			}
			child { [sibling distance=1cm]
				node (oc4) [oclass] {}
				child{
					node (oc10) [oclass] {}
				}
				child { 
					node (oc11) [oclass] {}
				}
			}
			child { 
				node (oc3) [oclass] {}
				child{
					node (oc7) [oclass] {}
					child{
						node (oc8) [oclassD1D2] {}
						child{
							node (oc9) [oclass] {}
						}
					}
				}
			}
			;
			
			\node at (0, -2.8) (e1) {$e_1$};
			\node at (0, -3.4) (e2) {$e_2$};
			\node at (-2, -3.4) (e3) {$e_3$};
			
			\draw[dashed, ->] (e1) -- (oc8);
			\draw[dashed, ->] (e1) -- (oc6);
			\draw[dashed, ->] (e2) -- (oc8);
			\draw[dashed, ->] (e3) -- (oc2);
			\end{tikzpicture}
		}
		
		\vspace{1em}
		
		\subfloat[][$\left\{e_1\right\} \leqa \left\{e_2, e_3\right\}$]{\centering
			\begin{tikzpicture}[
			oclass/.style={circle, draw=none, fill=black,
				text centered, anchor=north, text=white},
			oclassD1/.style={circle, draw=none, fill=lightgray,
				text centered, anchor=north, text=white},
			oclassD2/.style={circle, draw=none, fill=gray,
				text centered, anchor=north, text=white},
			oclassD1D2/.style={circle, draw=gray, line width=0.5mm, fill=lightgray,
				text centered, anchor=north, text=white},
			level distance=0.5cm, growth parent anchor=south
			]
			\node (oc1) [oclass] {} [<-, sibling distance=2cm]
			child{ [sibling distance=1cm]
				node (oc2) [oclassD2] {}
				child{
					node (oc5) [oclass] {}
				}
				child { 
					node (oc6) [oclass] {}
				}
			}
			child { [sibling distance=1cm]
				node (oc4) [oclassD2] {}
				child{
					node (oc10) [oclass] {}
				}
				child { 
					node (oc11) [oclassD1] {}
				}
			}
			child { 
				node (oc3) [oclass] {}
				child{
					node (oc7) [oclass] {}
					child{
						node (oc8) [oclass] {}
						child{
							node (oc9) [oclass] {}
						}
					}
				}
			}
			;
			
			\node at (0.5, -3.4) (e1) {$e_1$};
			\node at (0, -2.8) (e2) {$e_2$};
			\node at (-2, -3.4) (e3) {$e_3$};
			
			\draw[dashed, ->] (e1) -- (oc11);
			\draw[dashed, ->] (e2) -- (oc4);
			\draw[dashed, ->] (e3) -- (oc2);
			\end{tikzpicture}
		}
		\qquad
		\qquad
		\subfloat[][$\left\{e_1\right\} \leqa \left\{e_2\right\}$]{\centering
			\begin{tikzpicture}[
			oclass/.style={circle, draw=none, fill=black,
				text centered, anchor=north, text=white},
			oclassD1/.style={circle, draw=none, fill=lightgray,
				text centered, anchor=north, text=white},
			oclassD2/.style={circle, draw=none, fill=gray,
				text centered, anchor=north, text=white},
			oclassD1D2/.style={circle, draw=gray, line width=0.5mm, fill=lightgray,
				text centered, anchor=north, text=white},
			level distance=0.5cm, growth parent anchor=south
			]
			\node (oc1) [oclass] {} [<-, sibling distance=2cm]
			child{ [sibling distance=1cm]
				node (oc2) [oclassD2] {}
				child{
					node (oc5) [oclass] {}
				}
				child { 
					node (oc6) [oclass] {}
				}
			}
			child { [sibling distance=1cm]
				node (oc4) [oclassD2] {}
				child{
					node (oc10) [oclass] {}
				}
				child { 
					node (oc11) [oclassD1] {}
				}
			}
			child { 
				node (oc3) [oclass] {}
				child{
					node (oc7) [oclass] {}
					child{
						node (oc8) [oclass] {}
						child{
							node (oc9) [oclass] {}
						}
					}
				}
			}
			;
			
			\node at (0.5, -2.8) (e1) {$e_1$};
			\node at (-1, -3.4) (e2) {$e_2$};
			
			\draw[dashed, ->] (e1) -- (oc11);
			\draw[dashed, ->] (e2) .. controls (-0.2,-2.8) .. (oc4);
			\draw[dashed, ->] (e2) .. controls (-1.8,-2.8) .. (oc2);
			\end{tikzpicture}
		}
		
		\vspace{1em}
		
		\subfloat[][$\left\{e_1\right\} \not\leqa \left\{e_2\right\}; \ $ 
		$\left\{e_2\right\} \not\leqa \left\{e_1\right\}$]{\centering
			\begin{tikzpicture}[
			oclass/.style={circle, draw=none, fill=black,
				text centered, anchor=north, text=white},
			oclassD1/.style={circle, draw=none, fill=lightgray,
				text centered, anchor=north, text=white},
			oclassD2/.style={circle, draw=none, fill=gray,
				text centered, anchor=north, text=white},
			oclassD1D2/.style={circle, draw=gray, line width=0.5mm, fill=lightgray,
				text centered, anchor=north, text=white},
			level distance=0.5cm, growth parent anchor=south
			]
			\node (oc1) [oclass] {} [<-, sibling distance=2cm]
			child{ [sibling distance=1cm]
				node (oc2) [oclassD1] {}
				child{
					node (oc5) [oclass] {}
				}
				child { 
					node (oc6) [oclass] {}
				}
			}
			child { [sibling distance=1cm]
				node (oc4) [oclassD2] {}
				child{
					node (oc10) [oclassD1] {}
				}
				child { 
					node (oc11) [oclass] {}
				}
			}
			child { 
				node (oc3) [oclass] {}
				child{
					node (oc7) [oclass] {}
					child{
						node (oc8) [oclass] {}
						child{
							node (oc9) [oclass] {}
						}
					}
				}
			}
			;
			
			\node at (-1.5, -3.4) (e1) {$e_1$};
			\node at (0, -2.8) (e2) {$e_2$};
			
			\draw[dashed, ->] (e1) .. controls (-0.5,-2.8).. (oc10);
			\draw[dashed, ->] (e1) .. controls (-2,-2.8) .. (oc2);
			\draw[dashed, ->] (e2) -- (oc4);
			\end{tikzpicture}
		}
		\qquad
		\qquad
		\subfloat[][$\left\{e_1\right\} \sima \left\{e_2\right\}$]{\centering
			\begin{tikzpicture}[
			oclass/.style={circle, draw=none, fill=black,
				text centered, anchor=north, text=white},
			oclassD1/.style={circle, draw=none, fill=lightgray,
				text centered, anchor=north, text=white},
			oclassD2/.style={circle, draw=none, fill=gray,
				text centered, anchor=north, text=white},
			oclassD1D2/.style={circle, draw=gray, line width=0.5mm, fill=lightgray,
				text centered, anchor=north, text=white},
			level distance=0.5cm, growth parent anchor=south
			]
			\node (oc1) [oclass] {} [<-, sibling distance=2cm]
			child{ [sibling distance=1cm]
				node (oc2) [oclass] {}
				child{
					node (oc5) [oclass] {}
				}
				child { 
					node (oc6) [oclass] {}
				}
			}
			child { [sibling distance=1cm]
				node (oc4) [oclass] {}
				child{
					node (oc10) [oclass] {}
				}
				child { 
					node (oc11) [oclass] {}
				}
			}
			child { 
				node (oc3) [oclass] {}
				child{
					node (oc7) [oclass] {}
					child{
						node (oc8) [oclassD1D2] {}
						child{
							node (oc9) [oclass] {}
						}
					}
				}
			}
			;
			
			\node at (0.5, -3.4) (e1) {$e_1$};
			\node at (0.5, -2.6) (e2) {$e_2$};
			
			\draw[dashed, ->] (e1) -- (oc8);
			\draw[dashed, ->] (e2) -- (oc8);
			\end{tikzpicture}
		}
	\end{center}
	\caption{
		Examples of use cases of the preorder $\leqa$. 
		Circles represent ontology classes. 
		Solid arrows depict class subsumptions and dashed arrows depict class instantiations by individuals $e_1$, $e_2$, and $e_3$. 
		The light gray color identifies classes in $\msci(\ontology, e_1)$. 
		The dark gray color identifies classes in $\msci(\ontology, e_2)$ and $\msci(\ontology, e_3)$.
	}
	\label{fig:examples-leqa}
\end{figure}

\begin{example} Figure~\ref{fig:examples-leqa} depicts six examples for the application of $\leqa$:
	\begin{description}[style=multiline,leftmargin=0.7cm,font=\normalfont]
		\item[(a)] $\left\{e_1\right\}$ is more specific than $\left\{e_2, e_3\right\}$ even if $e_3$ instantiates a more specific class 
		than $e_1$, because of the more general individual $e_2$.
		\item[(b)] $\left\{e_1\right\}$ is more specific than $\left\{e_2, e_3\right\}$ since classes in $\msci(\ontology, e_1)$
		are either the same than those in $\msci(\ontology, e_2)$ or more specific than those in $\msci(\ontology, e_3)$.
		\item[(c)] $\left\{e_1\right\}$ is more specific than $\left\{e_2, e_3\right\}$ since the class in $\msci(\ontology, e_1)$ is more specific than the one in $\msci(\ontology, e_2)$. There is no need to compare it with the class in $\msci(\ontology, e_3)$. 
		\item[(d)] This example, similar to (c), illustrates the occurrence of the same behavior regardless of classes being instantiated by a single or by several individuals.
		\item[(e)] $\left\{e_1\right\}$ and $\left\{e_2\right\}$ cannot be compared. Unlike the two latter examples, here, $e_1$ instantiates a class that is more specific than the class instantiated by $e_2$, but also a class that is not comparable.
		\item[(f)] $\left\{e_1\right\}$ and $\left\{e_2\right\}$ are equivalent by instantiating the same most specific class.
	\end{description}
\end{example}

\section{Application to Pharmacogenomic Knowledge (details)}
\label{section:details-experimentation}

We experimented our methodology with PGxLOD\footnote{\url{https://pgxlod.loria.fr}}~\cite{monnin2018pgxo}, a PGx knowledge base represented in the $\mathcal{ALHI}$ Description Logic~\cite{baader2003}.

In PGxLOD, PGx tuples are represented using classes and predicates of the PGxO ontology\footnote{\url{https://pgxo.loria.fr}}. 
PGx tuples are $n$-ary, and thus, they are reified as instances of the \texttt{Pharmaco\-genomic\-Relationship} class.
All the individuals involved in PGx tuples instantiate the \texttt{Drug}, \texttt{Genetic\-Factor}, or \texttt{Phenotype} classes.
They are linked with reified PGx tuples by predicates qualifying their association to tuples. 
These predicates are organized in a hierarchy defined by subsumption axioms, such as $\mtt{causes} \sqsubseteq \mtt{influences} \sqsubseteq \mtt{isAssociatedWith}$.
It is noteworthy that, in PGxLOD, \texttt{partOf} links indicate that instances of \texttt{GeneticFactor} compose others such instances.
For example, a genomic variation may be part of a gene.
Similarly, instances of \texttt{Phenotype} may have dependencies, expressed with \texttt{dependsOn} links.
These dependencies enable representing complex phenotypes that refer to other phenotypes or drugs.
For example \textit{warfarin-caused hemorrhage} is a phenotype linked with \texttt{dependsOn} to \textit{hemorrhage} and \textit{warfarin}.
The TBox of PGxLOD contains, alongside PGxO, three other ontologies: individuals representing drugs may instantiate classes from ATC or ChEBI, and individuals representing phenotypes may instantiate classes from MeSH.
Table~\ref{table:pgxlod-statistics} provides global statistics about PGxLOD.

\begin{table}
	\caption{Statistics about PGxLOD. \# denotes ``number of''. 
		Instances linked by \texttt{owl:sameAs} are counted separately. \texttt{partOf} links are counted without transitivity inference.
		All PGx tuples were programmatically extracted from their sources, except the ten tuples from EHRs that were manually added as a proof of concept.
	}
	\label{table:pgxlod-statistics}
	\begin{center}
		\begin{tabular}{lrr||rp{2cm}r}
			\toprule
			Class & \# instances & & & Predicate &  \# links \\
			\midrule
			\texttt{Drug} & 47,584 & & &\texttt{partOf} & 16,697 \\
			\texttt{GeneticFactor} & 464,302 & & & \texttt{dependsOn} & 23,976 \\
			\texttt{Phenotype} & 61,330 & & & &  \\
			\texttt{PharmacogenomicRelationship} & 50,435 & & & &  \\
			\ \ \rotatebox[origin=c]{180}{$\Lsh$} From PharmGKB (structured data) & 3,650 & & & &  \\
			\ \ \rotatebox[origin=c]{180}{$\Lsh$} From PharmGKB (clinical annotations) & 10,240 & & & &  \\
			\ \ \rotatebox[origin=c]{180}{$\Lsh$} From biomedical literature & 36,535 & & & &  \\
			\ \ \rotatebox[origin=c]{180}{$\Lsh$} From EHRs & 10 & & & &  \\
			\bottomrule
		\end{tabular}
	\end{center}
\end{table}

To apply the matching rules on PGx tuples, we specified their arguments.
Each argument of a tuple is the set of individuals with a specific type (\texttt{Drug}, \texttt{GeneticFactor}, or \texttt{Phenotype}) that are linked with a specific predicate to the tuple.
For example, given $pgt$ a PGx tuple, $\pi_{\texttt{Phenotype},\texttt{causes}}(pgt)$ contains all the phenotypes caused by $pgt$.
Hence, as there are 3 types of individuals and 38 predicates, PGx tuples have $3 \times 38 = 114$ arguments.
Once arguments of tuples are specified, their associated preorders can be
defined. 
Based on the available data and knowledge in PGxLOD, it makes 
sense to use the $\leqp{partOf}$ preorder for arguments involving instances of 
\texttt{GeneticFactor}. 
Similarly, we use $\leqd$ and $\leqop$ as preorders for 
arguments respectively involving instances of \texttt{Drug} and \texttt{Phenotype}, where $\ontologyd$ is the concatenation of ATC and ChEBI, and $\ontologyp$ is the MeSH ontology.

Finally, to apply Rule~\ref{matching-rule:5}, a natural three-way partition of arguments appears based on the three types of involved individuals.
Therefore, discarding predicates, we gather all drugs, genetic factors, and phenotypes involved in a tuple in three aggregated arguments.
To benefit from dependencies of complex phenotypes, we choose to add them to the aggregated arguments corresponding to their type.
For example, in \textit{warfarin-caused hemorrhage}, \textit{hemorrhage} is added to the aggregated argument representing phenotypes and \textit{warfarin} is added to the one representing drugs. We arbitrarily set $\gamma_{\neq\Delta} = 3$, $\gamma_S = 0.8$, and $\gamma_C = 2$.
These values mean that two PGx tuples $pgt_1$ and $pgt_2$ will be matched by Rule~\ref{matching-rule:5} if their three aggregated arguments are specified (\textit{i.e.}, different from $\Delta$).
Additionally, each of the three aggregated arguments of $pgt_1$ must have at least 80\% of comparable individuals with the same aggregated argument of $pgt_2$, or at least two aggregated arguments of $pgt_1$ must be comparable with the same aggregated arguments of $pgt_2$.

To illustrate the interest of this formalization as well as reasoning mechanisms from Description Logics, let $pgt_1$ and $pgt_2$ be two PGx tuples. 
$pgt_1$ causes a phenotype $ph_1$ and is associated with a phenotype $ph_2$, and $pgt_2$ is associated with both phenotypes. 
Thus, by applying reasoning mechanisms along the hierarchy of predicates, it follows that:
\[\pi_{\texttt{Phenotype},\texttt{causes}}(pgt_1) = \big\{ph_1\big\}; \ \pi_{\texttt{Phenotype},\texttt{isAssociatedWith}}(pgt_1) = \big\{ph_1,\ ph_2\big\}\]
\[\pi_{\texttt{Phenotype},\texttt{causes}}(pgt_2) = \Delta; \  \pi_{\texttt{Phenotype},\texttt{isAssociatedWith}}(pgt_2) = \big\{ph_1,\ ph_2\big\}\]
\noindent By definition of $\leqop$, $\pi_{\texttt{Phenotype},\texttt{causes}}(pgt_1) \leqop \pi_{\texttt{Phenotype},\texttt{causes}}(pgt_2)$ as well as $\pi_{\texttt{Phenotype},\texttt{isAssociatedWith}}(pgt_1) \leqop \pi_{\texttt{Phenotype},\texttt{isAssociatedWith}}(pgt_2)$. 
Therefore, by applying Rule~\ref{matching-rule:3}, 
$pgt_1$ is more specific than $pgt_2$. 
This makes sense as the predicate connecting $ph_1$ with $pgt_1$ is more specific with than the one used with $pgt_2$.

We implemented our matching methodology in C++ with multithreading. 
Our code is available on GitHub\footnote{\url{https://github.com/pmonnin/tcn3r}}. 
Our program interacts with the knowledge base thanks to SPARQL queries. Previously, we indicated that an unspecified argument of an $n$-ary tuple is set to $\Delta$. 
Accordingly, when a SPARQL query returns $\emptyset$ for an argument of a tuple, it is interpreted as if it is returning $\Delta$.
On PGxLOD, the matching rules led to perform $\binom {50,435}{2} = 1,271,819,395$ comparisons in approximately 54 hours using 4 cores and 15 GB of RAM.

\section{Discussion (details)}
\label{section:details-discussion}

By looking more closely at the results in Table~\ref{table:results}, we notice that all \texttt{owl:sameAs} links are intra-source and thus indicate duplicates.
This is expected in the case of the literature since several articles could mention the same tuple.
The 5 \texttt{skos:closeMatch} links between tuples from structured data and clinical annotations of PharmGKB highlight expected agreements between these two related sources.
However, linked tuples are expressed with different individuals instantiating the same ontology classes, preventing their reconciliation with \texttt{owl:sameAs}.

The results of Rule~\ref{matching-rule:4} underline that sources may contain tuples with comparable arguments.
Source owners can benefit from such results by considering adding a tuple formed by the most specific arguments of the matched tuples.
We notice that only Rule~\ref{matching-rule:5} generates links between the tuples from EHRs and other sources.
As tuples from EHRs are manually represented, there are only a few of them, minimizing the chance of overlap with other sources.
Additionally, phenotypes involved in tuples from EHRs are very specific, making their comparison with phenotypes from biomedical literature or PharmGKB difficult.

Regarding our method, we believe our rules are simple and abstract enough to be useful in other domains.
Additionally, their readability facilitates their review by experts, for example, to define the arguments and preorders to use in another application domain. 
Finally, it is noteworthy that the $\leqa$ preorder may result in many equivalences if the ontology $\ontology$ is not granular enough in terms of width and depth. 
Such equivalences may make sense, depending on the application domain and the ontology.
If not, the two other preorders (\textit{i.e.}, $\subseteq$ and $\leqp{p}$) could be used.
It is for now up to experts to choose the correct preorder for each argument.
However, in future works, we could investigate metrics about domain knowledge that may guide their choice.

\end{document}